%% file: tcl_paper.tex
\documentclass{article}

\usepackage{microtype}
\usepackage{graphicx}
\usepackage{subfigure}
\usepackage{booktabs} % for professional tables
\usepackage{layouts}
\usepackage{pgf}
\usepackage{tikz}
\usepackage[utf8]{inputenc}
\usepackage{pgfplots}
\DeclareUnicodeCharacter{2212}{−}
\usepgfplotslibrary{groupplots,dateplot}
\usetikzlibrary{patterns,shapes.arrows}
\pgfplotsset{compat=newest}
\usepackage{adjustbox}
\usepackage{array}
\usepackage{xcolor}
\usepackage{soul}
\usepackage{appendix}
\usepackage{listings}
\usepackage{amsmath,amssymb,amsthm,amsfonts}
\usepackage{mathtools}
\usepackage{booktabs}

\usepackage{hyperref}

\definecolor{deepred}{rgb}{0.6,0,0}
\definecolor{codeblue}{rgb}{0.25,0.5,0.5}
\definecolor{keywords}{rgb}{0.08,0.54,.02}
\newcommand\Tau{\mathcal{T}}
\newcommand\Beta{\mathcal{B}}%
\newcommand\Normal{\mathcal{N}}%
\newcommand\Dataset[1]{\mathcal{D}_\textrm{#1}}%
\newcommand\Loss{\mathcal{L}}%
\newcommand\Algo{\mathcal{A}}%
\DeclareMathOperator{\E}{\mathbb{E}}
\DeclarePairedDelimiterX{\infdivx}[2]{(}{)}{%
  #1\;\delimsize\|\;#2%
}

\DeclareMathOperator*{\argmax}{arg\,max}

\newcommand{\norm}[1]{\left\lVert#1\right\rVert}
\newcolumntype{R}[2]{%
    >{\adjustbox{angle=#1,lap=\width-(#2)}\bgroup}%
    l%
    <{\egroup}%
}
\usepackage[accepted]{icml2021}

\icmltitlerunning{Trajectory Contrastive Learning}

\begin{document}

\twocolumn[
\icmltitle{Improving Context-Based Meta-Reinforcement Learning \\
with Self-Supervised Trajectory Contrastive Learning}



\icmlsetsymbol{equal}{*}

\begin{icmlauthorlist}
\icmlauthor{Bernie Wang}{equal,berk}
\icmlauthor{Simon Xu}{equal,berk}
\icmlauthor{Kurt Keutzer}{berk}
\icmlauthor{Yang Gao}{tsing}
\icmlauthor{Bichen Wu}{fb}
\end{icmlauthorlist}

\icmlaffiliation{berk}{University of California, Berkeley}
\icmlaffiliation{tsing}{Tsinghua University}
\icmlaffiliation{fb}{Facebook Research}

\icmlcorrespondingauthor{Bernie Wang}{berniewang@berkeley.edu}
\icmlcorrespondingauthor{Simon Xu}{simon0xzx@gmail.com}

\icmlkeywords{Machine Learning, ICML}

\vskip 0.3in
]



\printAffiliationsAndNotice{\icmlEqualContribution} % otherwise use the standard text.

\begin{abstract}
Meta-reinforcement learning typically requires orders of magnitude more samples than single task reinforcement learning methods. This is because meta-training needs to deal with more diverse distributions and train extra components such as context encoders. To address this, we propose a novel self-supervised learning task, which we named Trajectory Contrastive Learning (TCL), to improve meta-training. TCL adopts contrastive learning and trains a context encoder to predict whether two transition windows are sampled from the same trajectory. TCL leverages the natural hierarchical structure of context-based meta-RL and makes minimal assumptions, allowing it to be generally applicable to context-based meta-RL algorithms. It accelerates the training of context encoders and improves meta-training overall. Experiments show that TCL performs better or comparably than a strong meta-RL baseline in most of the environments on both meta-RL MuJoCo (5 of 6) and Meta-World benchmarks (44 out of 50).
\end{abstract}

\input{1-intro}
\input{2-related-work}
\input{3-problem-statement}

\input{4-method}

\input{5-experiments}
\input{6-conclusion}

\newpage
\bibliography{tcl_paper}
\bibliographystyle{icml2021}
\input{7-appendix}
\end{document}

%% file: 1-intro.tex
\section{Introduction}
\label{intro}

\begin{figure}[t!]
%\begin{center}
\includegraphics[width=0.9\linewidth]{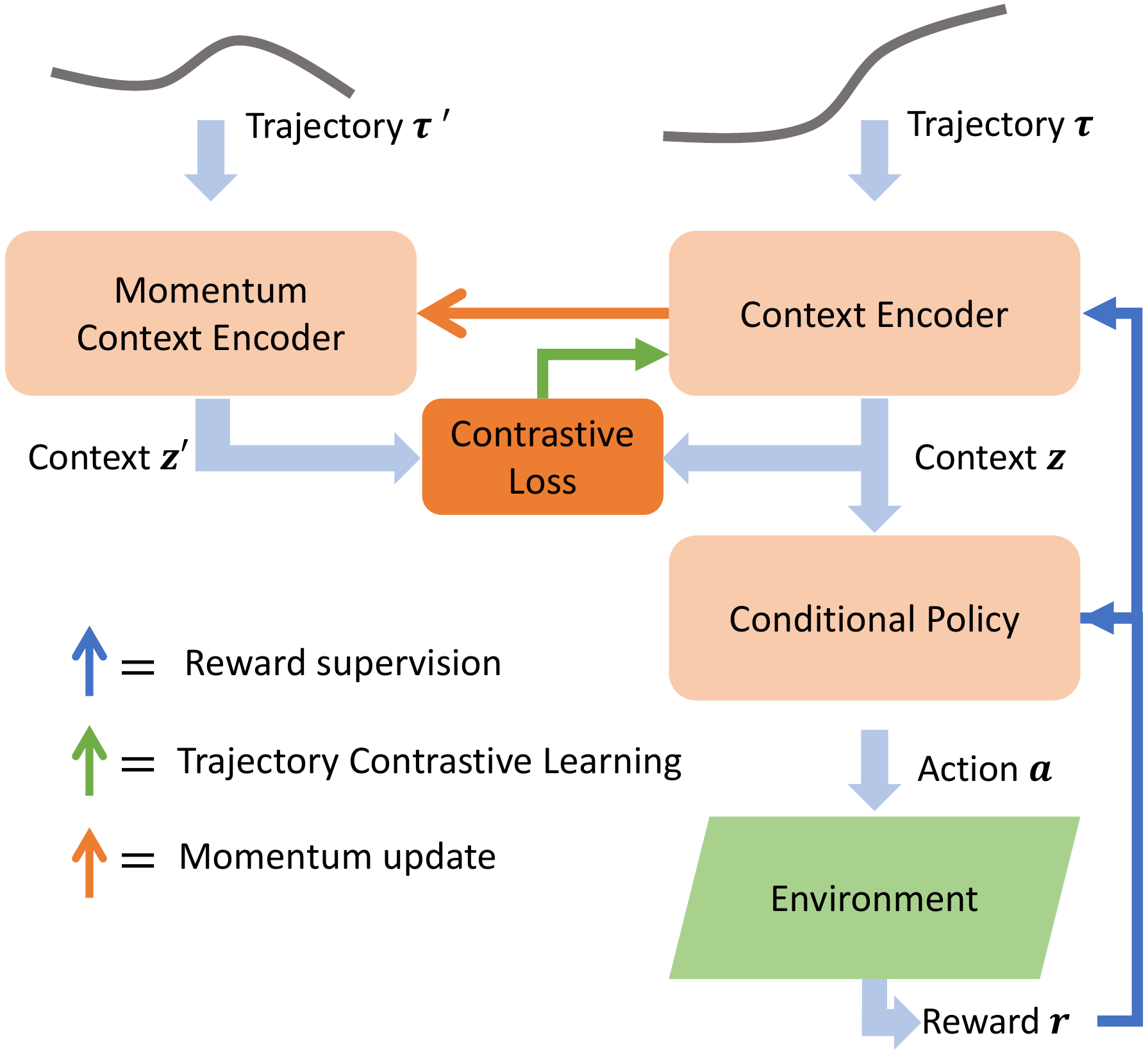}
\vskip -.3cm
  \caption{
  Overview of our TCL method in a generic context-based meta-RL framework. A context-based meta-RL method has two components, a context encoder and a conditional policy. The context encoder takes in exploratory trajectories and outputs context variable $\mathbf{z}$ denoting current task properties. The conditional policy takes in states and conditions on $\mathbf{z}$ to output actions. Previous meta-RL methods only rely on the sparse task rewards to train both the context encoder and the policy. However, due to the weak supervision and many components to train at the same time, previous methods suffer from unstable training and inferior performances. TCL improves this using a contrastive auxiliary loss that directly provides strong supervision to the context encoder.}
\label{fig:concept}
\vskip -.7cm
\end{figure}

Although reinforcement learning (RL) has achieved impressive successes in a number of fields, such as Go~\citep{silver2016mastering}, robotic manipulation~\citep{andrychowicz2020learning} and game playing~\citep{OpenAI_dota}, it usually needs millions or even billions of environment interactions to learn those tasks. In contrast, humans are able to acquire a new skill very fast based on their prior experiences. Meta-RL aims to bridge this gap by training on a distribution of tasks, and quickly adapts to a new task during test time. Recent meta-learning algorithms~\citep{finn2017modelagnostic,duan2016rl2,rakelly2019efficient,fakoor2020metaqlearning} can quickly adapt on classification, regression, and policy learning problems. 

Meta-RL saves millions of samples for each new task during test time. However, training meta-RL is much more challenging than single-task RL, as the training requires orders of magnitudes more samples and may yield worse performance compared with single-task RL. This is not surprising, as during meta-training, an agent not only has to learn to infer the characteristics of each task, such as the dynamics and rewards, but also has to learn the corresponding policies. Meanwhile, recent meta-RL algorithms leverage a context encoder in addition to the policy, but they train both context encoder and policy in a naive end-to-end fashion, relying on the task reward as the only supervision. This limits the performance and applicability of meta-RL.

In this paper, we propose trajectory contrastive learning (TCL) to improve meta-training (Figure \ref{fig:concept}). During meta-training, the agent's trajectories of environmental interactions are collected and stored in a replay buffer. On top of the original meta-training pipeline, we additionally train the context encoder to perform TCL, a proxy learning task that asks the context encoder to predict whether two transition windows are cropped from the same trajectory. Since trajectories can reveal characteristics about the task, such as dynamics and rewards, this proxy allows the encoder to learn meaningful representations about the underlying tasks. Following the previous success of contrastive learning in RL, we adopt the key-query architecture proposed in CURL \cite{laskin_srinivas2020curl} and train the encoder with the InfoNCE loss \cite{oord2018representation}. Since the pseudo-label needed for this task is freely available when collecting trajectories, TCL does not require additional labels. TCL naturally arises from the structure of the meta-RL and makes no extra assumptions on tasks and algorithms, allowing it to be widely applicable to context-based meta-RL algorithms. 

We conduct experiments to thoroughly study how TCL performs under a wide range of settings. Combined with PEARL \cite{rakelly2019efficient},  a strong meta-RL baseline, TCL-PEARL outperforms PEARL in 5 out of 6 environments on the widely used MuJoCo~\citep{todorov2012mujoco} meta-RL benchmark. However, we find that this baseline performance is too close to the oracle goal-conditioned RL due to relatively simple task distributions. We further benchmark on Meta-World~\citep{yu2020meta}, a more challenging benchmark consists of 50 environments, and again we perform better than (35) or comparably with (9) the baseline in 44 out of 50 environments.  

%% file: 2-related-work.tex
\section{Related Work}

\textbf{Meta-Learning.} 
The classical meta-learning formulation is  based on the idea of learning a parameter initialization that can quickly be optimized for a given task. Early work \cite{Baxter_1995, schmidhuber1987evolutionary, thrun_pratt} proposed having a meta-learner learn from a collection of base learners that have each been optimized for a particular task. \citet{hochreiter, andrychowicz2016learning, li2016learning} applied this idea to learning to optimize deep neural networks. Meta-learning has shown success in few-shot supervised learning tasks like image classification \cite{finn2017modelagnostic, snell2017prototypical, vinyals2017matching} and generative modeling \cite{rezende2016oneshot, edwards2017neural}. In RL, policies \cite{duan2016rl2, finn2017modelagnostic, mishra2017simple, rakelly2019efficient, fakoor2020metaqlearning} and dynamics models \cite{saemundsson2018meta, nagabandi2019learning} have been meta-trained to generalize to unseen tasks. 

Gradient-based meta-RL algorithms \cite{finn2017modelagnostic, rothfuss2018promp, xu2018learning, stadie2019considerations} learn a policy initialization that can attain single-task level performance on a new task after one or few policy gradient steps. %However, they suffer from poor meta-training and meta-testing sample efficiency because they are on-policy and leverage gradient-based adaptations. 
Context-based meta-RL methods learn from past experience by leveraging context, which is formulated as a collection of past agent-environment interactions~\citep{duan2016rl2, wang2016learning, fakoor2020metaqlearning,rakelly2019efficient}. Context-based meta-RL trains both the context encoder and policy in a naive end-to-end fashion, and the context encoder is trained based on the reward signal's indirect supervision. We find that this is insufficient and can lead to worse performance and sample efficiency. Our TCL method provides direct supervision to the context encoder and thus improves meta-training.

\textbf{Auxiliary Tasks in RL}
Auxiliary tasks have been used to improve RL performance. Prior work that used auxiliary self-supervised tasks include using future prediction \cite{oord2018representation, Jaderberg_2019} or reconstruction \cite{shelhamer2017loss} to improve sample efficiency and performance of end-to-end RL. Contrastive learning has been used in RL to extract reward functions \cite{sermanet2018timecontrastive} and learn representations of visual inputs \cite{dwibedi2019learning, laskin_srinivas2020curl}. Our method is also using an auxiliary loss to improve performance. Unlike previous work which aims to improve RL algorithm performance, we deal with meta-learning problems and aim to learn representations of tasks rather than state observations. Concurrent and independent to our work, \citet{fu2020towards} also explores using contrastive learning to improve the sample efficiency of the training and exploration of meta-learning algorithms. Different from \citet{fu2020towards}, our work is based on probabilistic context encoders, and the contrastive learning is in a probabilistic metric space.

\textbf{Contrastive Learning.}
Contrastive learning is an unsupervised learning method that aims to learn rich representations of high dimensional data by enforcing similar pairs in the input space to also be similar in the representation space. Early work \cite{dosovitskiy2015discriminative} proposed instance discrimination, which defines similar pairs as augmentations of the same instance and dissimilar pairs are those of different instances. \citet{wu2018unsupervised} proposed using a memory bank to store instance representations and reformulated instance discrimination as a dictionary lookup task where a query encoding should match its key encoding with respect to a set of keys. Positive query-key pairs are encodings of different views of the same instance while negative pairs are of different instances. More recently, contrastive learning methods like CPC \cite{henaff2020dataefficient}, MoCo \cite{he2020momentum}, SimCLR \cite{chen2020simple}, SimSiam \cite{chen2020exploring} and BYOL \cite{grill2020bootstrap} have shown a lot of successes in computer vision. TCL adopts the query-key architecture of \citet{he2020momentum}, but more closely follows CURL \cite{laskin_srinivas2020curl} where queries and keys are windows of the trajectories. 

%% file: 3-problem-statement.tex
\section{Problem Statement}
In this section, we formally define the meta-RL problem and introduce the objective of the paper. 

Meta-RL aims to train an agent that is able to tackle a set of tasks $\{\mathcal{T}\}$ drawn from a distribution $\rho(\Tau)$. Each task is modeled as a Markov Decision Process (MDP) as
\begin{equation}
    \Tau = (p(\textbf{s}_0), p(\textbf{s}_{t+1}|\textbf{s}_t, \textbf{a}_t), r(\textbf{s}_t, \textbf{a}_t), \gamma, T),
\end{equation}
where $\gamma \in [0, 1]$, $\textbf{a}_t \in A \subset \mathbb{R}^d$, $\textbf{s}_t \in S \subset \mathbb{R}^p$. Here, $A$ is the action space, $S$ is the state space, $p(\textbf{s}_0)$ is the initial state distribution, $p(\textbf{s}_{t+1}|\textbf{s}_t, \textbf{a}_t)$ is the transition distribution, $r(\textbf{s}_t, \textbf{a}_t)$ is the reward function, $\gamma$ is the discount factor, and $T$ is the time horizon. At timestep $t$, an agent takes an action $\textbf{a}_t$ sampled from policy distribution $\pi_{\theta}(\textbf{a}_t|\textbf{s}_t)$ and transitions to the next state $\textbf{s}_{t+1}$ sampled from $p(\textbf{s}_{t+1}|\textbf{s}_t, \textbf{a}_t)$, collecting scalar reward $r(\textbf{s}_t, \textbf{a}_t)$.  We assume all tasks in $\rho(\Tau)$ share the same $S$, $A$, $\gamma$, $T$, but different tasks have different  $p(\textbf{s}_{t+1}|\textbf{s}_t, \textbf{a}_t)$ and  $r(\textbf{s}_t, \textbf{a}_t)$.

The goal of meta-RL is to maximize the expected reward
\begin{equation}
    \E_{\Tau \in \rho(\Tau)}[\E_{\tau \sim \pi_{\theta(\Tau)}}[\textstyle{\sum_{t=0}^{T-1}{\gamma^{t}r(\textbf{s}_t, \textbf{a}_t)}}]],
\end{equation}
where the inner expectation is taken on the trajectories sampled with a policy $\pi_{\theta(\Tau)}$ adapted to task $\Tau$, and the outer expectation is taken on the task distribution. 

During the testing stage, we assume that the state transition and reward are unknown to the agent. The agent is given the opportunity to interact with the environment to infer about the task. The interactions are recorded as a sequence of transition tuples, which we define as a trajectory:
\begin{equation}
\label{eqn:trajectory}
    \tau = \{\tau_t | \tau_t = (\textbf{s}_t, \textbf{a}_t, r_t, \textbf{s}_t^\prime)\}_{t=1:T},
  \end{equation}
where $\textbf{s}_t^\prime = \textbf{s}_{t+1}$, $\tau_t$ is a transition tuple. 

Meta-RL algorithms are trained on a set of training tasks $\Dataset{train}$ and evaluated on a held-out set of test tasks $\Dataset{test}$. The meta-training procedure optimizes the policy on a set of training tasks $\Dataset{train}$ such that 
\begin{equation}
    \theta_\textrm{meta} = \argmax_{\theta} \E_{\Tau \in \Dataset{train}}[l_{\textrm{meta}}^\Tau(\theta)],
\end{equation}
where $l_\textrm{meta}^\Tau(\theta)$ is the algorithm-dependent loss for task $\Tau$. 

The training of meta-RL is more challenging than single-task RL, usually requiring orders of magnitude more samples and yielding worse performance. This is due to the fact that meta-training needs to deal with more diverse distributions and that there are more algorithm components in meta-RL that need to be trained. 
Our goal is to improve the training of context-based meta-RL, which can lead to better sample efficiency and task performance.

%% file: 4-method.tex
\section{Method}
\label{sec:method}
\subsection{Context-Based Meta-RL}

Recent work on meta-RL \cite{rakelly2019efficient} decomposes the meta-RL solution to two parts: a context encoder that infers the task context and a conditional policy based on the context. Before performing a task, an agent is given the opportunity to explore the environment and collect a set of trajectories $\{\tau_{i,t}\}_{i=1:N, t=1:T}$, with $\tau_{i,t}$ denoting the $t$-th transition in the $i$-th trajectory. The trajectories are used as a context to feed into the context encoder $f_\phi(\{\tau_{i,t}\})$. The context encoder is parameterized by $\phi$ and computes a context embedding (or task embedding) $\textbf{z}$, on which the policy, parameterized by $\theta$, is conditioned as $\pi_\theta(\textbf{a} | \textbf{s}, \textbf{z})$. 
Following \citet{rakelly2019efficient}, a context encoder is instantiated as an inference network, and the task embedding is modeled as a probabilistic latent variable $\textbf{z}$. We use the context encoder to estimate the posterior $p(\textbf{z} | \{\tau_{i,t}\})$ as 
\begin{equation}
\label{eqn:encoder}
    f_\phi(\textbf{z} | \{\tau_{i,t}\}) \propto \prod_{\substack{i=1:N \\ t=1:T}} \Normal(f^\mu_\phi(\tau_{i,t}), f^\sigma_\phi(\tau_{i,t}))
\end{equation}
Given context in the form of a batch of transitions, the encoder $f_\phi$, implemented as a multi-headed MLP, infers a batch of Gaussian factors parameterized by the mean $f^\mu_\phi(\tau_{i,t})$ and standard deviation $f^\sigma_\phi(\tau_{i,t})$. The task embedding $\textbf{z}$ is the product of independent factors.

\subsection{Trajectory Contrastive Learning}
\label{sec:contrastive_learning}

\begin{figure}[!htb]
\includegraphics[width=1\linewidth]{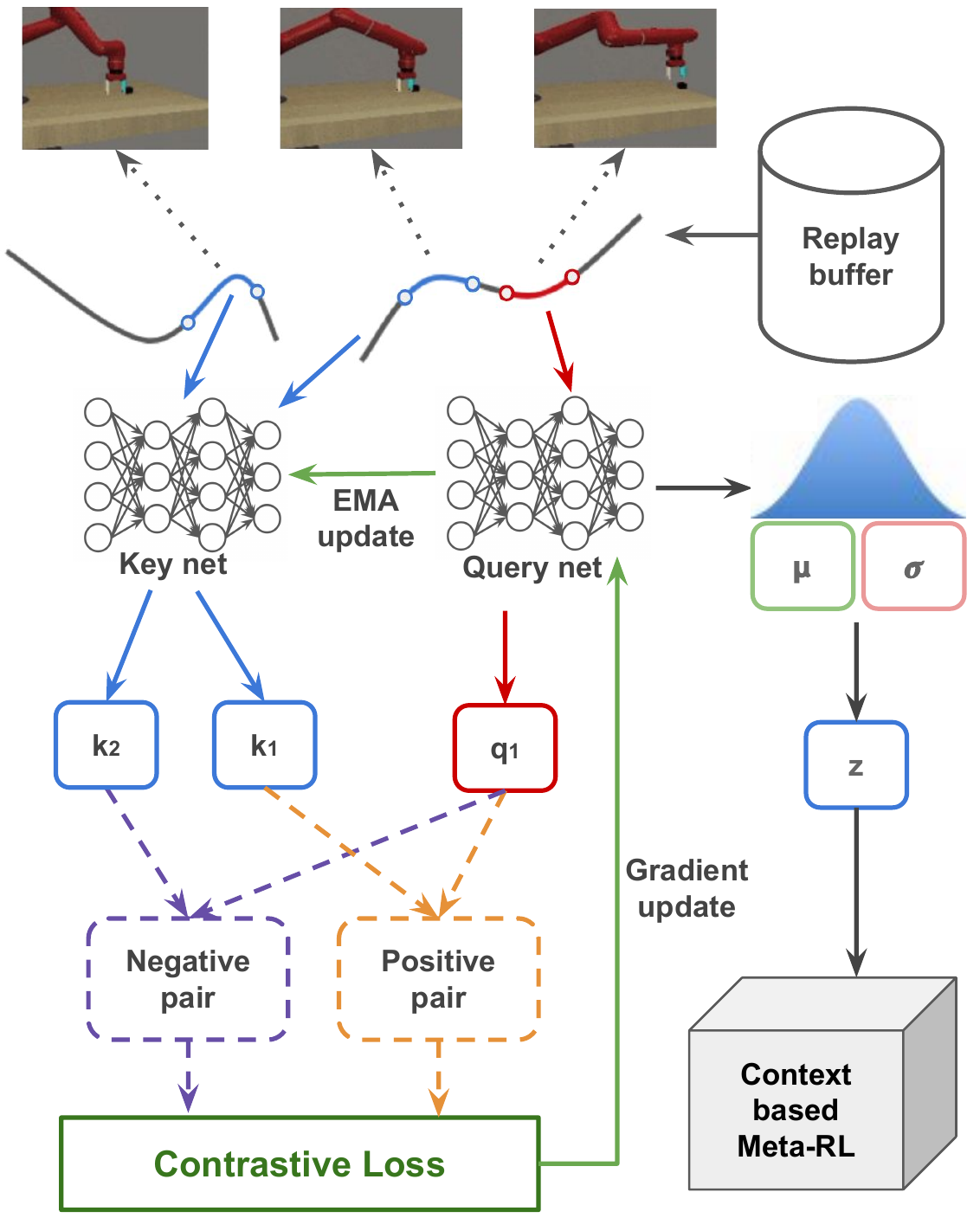}
\vskip -.3cm
    \caption{\textbf{Trajectory contrastive learning.} On top of the regular meta-training, we sample trajectories from a replay buffer and crop transition windows from trajectories. Windows are passed through a key and a query context encoder, and the output embeddings are compared with each other. Windows from the same trajectory are regarded as a positive pair, while those from different trajectories are regarded as negative pairs. We use a contrastive loss to train the positive pairs to be more similar to each other, and the negative pairs are trained to be dissimilar to each other. TCL is performed on top of the original meta-training and is generally applicable to context-based meta-RL algorithms.
}
\label{fig:tcl_architecture}
\end{figure}

In \citet{rakelly2019efficient}, $f_\phi(\cdot)$ is trained end-to-end together with the context conditioned policy $\pi_\theta(\textbf{a} | \textbf{s}, \textbf{z})$, relying on the reward as the only supervision.
In this work, we propose Trajectory Contrastive Learining (TCL) to improve context-based meta-RL training. On top of the original meta-training pipeline, we additionally train the context encoder to perform TCL, a proxy learning task that asks the encoder to predict whether two transition windows are cropped from the same trajectory. Since trajectories reveal characteristics about the tasks, such as dynamics and rewards, this proxy trains the encoder to develop meaningful representations for the underlying tasks. Figure \ref{fig:tcl_architecture} illustrates the idea of TCL.

\subsubsection{Proxy task for trajectory contrastive learning}
TCL is a task that asks a context encoder  $f_\phi(\cdot)$ to predict whether two windows of transitions are cropped from the same trajectory. The label for this task can be obtained for free. Meta-RL operates on trajectories corresponding to different tasks. During meta-training, we collect trajectories corresponding to different tasks and store them in a replay buffer. Based on this, we can obtain the information about which trajectory a transition window is sampled from, and leverage this as pseudo-labels for our proxy task. More specifically, from a replay buffer that contains a set of trajectories $\{\tau_{i,t}\}_{i=1:N, t=1:T}$, we randomly crop trajectory windows and form a set 
\begin{equation}
    \{w_{i,t} | w_{i,t} = \tau_{i, t:t+W}\},
\end{equation}
where each $w_{i,t}$ is a window cropped from trajectory-$i$, starting from time step-$t$, and contains $W$ transitions. We treat windows $w_{i,t}$ and $w_{i,t'}$ as a positive pair if they come from the same trajectory $i$, though their starting times are different. We treat windows $w_{i,t}$ and $w_{j,t'}$ as negative pairs if $i \neq j$, indicating that they come from different trajectories. 

Our proxy task follows the instance discrimination task in CURL \cite{laskin_srinivas2020curl}: a query matches a key if they are encoded augmentations of the same instance. This can be understood as a dictionary lookup task where a query encoding should match its key encoding with respect to a set of keys. Positive query-key pairs are encodings of different views of the same instance while negative pairs are encoded views of different instances. In TCL, we generate a set of queries $Q = \{q_i\}_{i \in [0, N-1]}$ and set of keys $K = \{k_i\}_{i \in [0, N-1]}$ from $N$ trajectories such that $q_i = f_{\phi_q}(w_{i,t})$ and $k_i = f_{\phi_k}(w_{i,t'})$. The set of keys $K$ can be partitioned for a particular query $q_i$ such that the corresponding positive key is $k_i$ and the negative keys are $k \in K \setminus \{k_i\}$. The proxy task is for every query $q_i$, maximize its agreement with its positive key $k_i$ while minimizing its agreement with each negative key $k \in K \setminus \{k_i\}$. We follow the momentum encoding procedure proposed in MoCo \cite{he2020momentum} by using a query encoder $f_{\phi_q}$, parameterized by $\phi_q$, for encoding queries and a key encoder $f_{\phi_k}$, parameterized by $\phi_k$, for encoding keys. The query encoder is trained with gradient updates and the key encoder's parameters $\phi_k$ are updated through exponential moving average of query parameter  $\phi_q$. We refer readers to \cite{he2020momentum, laskin_srinivas2020curl} for details.

\subsubsection{Trajectory Contrastive Loss}
\begin{algorithm}[tb]
   \caption{Trajectory Contrastive Loss}
   \label{alg:tcl_contrastive_loss}
\begin{algorithmic}[1]
    \FUNCTION{TCL(queries \textbf{Q}, keys \textbf{K})}
    \STATE $\textbf{D}_\mu$ = PairwiseSquaredL2Distance($\textbf{Q}_\mu$, $\textbf{K}_\mu$)
    \STATE $\textbf{D}_\sigma$ = PairwiseSquaredL2Distance($\textbf{Q}_\sigma$, $\textbf{K}_\sigma$)
    \STATE scores = $ - (\textbf{D}_\mu + \textbf{D}_\sigma)$
    \STATE logits = scores - max(scores, axis=1)
    \STATE labels = arange(logits.shape[0]) 
    \STATE \textbf{return} CrossEntropyLoss(logits, labels)
    \ENDFUNCTION
    
    %\FUNCTION{PairwiseSquaredL2Distance(\textbf{A}, \textbf{B})}
    %\STATE \textbf{C} = sum(square(\textbf{A}), axis=1)
    %\STATE \textbf{D} = sum(square(\textbf{B}), axis=1).T
    %\STATE \textbf{return} \textbf{C} + \textbf{D} - $2 \textbf{A} \textbf{B}^T$
    %\ENDFUNCTION
\end{algorithmic}
\end{algorithm}
To conduct contrastive learning, we need to define the similarity metrics between query-key pairs computed by the context encoder. Since our trajectory representations are modeled as probabilistic latent variables, directly computing the similarity of sampled embedding vectors can be very noisy. Instead, we consider metrics that express similarities between distributions. Following Equation (\ref{eqn:encoder}), for a query-key pair represented as normal distributions 
% $q \propto \Normal(\boldsymbol{\mu}_1, \boldsymbol{\sigma}_1)$ and $k \propto \Normal(\boldsymbol{\mu}_2, \boldsymbol{\sigma}_2)$. 
$q \sim \Normal(f^\mu_{\phi_q}(w_{i,t}), f^\sigma_{\phi_q}(w_{i,t}))$ and $k \sim \Normal(f^\mu_{\phi_k}(w_{i',t'}), f^\sigma_{\phi_k}(w_{i',t'}))$, we compute the similarity function based on the negative Wasserstein distance as
% \begin{equation}
%     \textrm{sim}(q, k) = -\norm{\boldsymbol{\mu}_1 - \boldsymbol{\mu}_2}^2_2 - \norm{\boldsymbol{\sigma}_1 - \boldsymbol{\sigma}_2}^2_2.
% \end{equation}
\begin{equation}
    \begin{split}
    \textrm{sim}(q, k) &= -\left(\norm{f^\mu_{\phi_q}(w_{i,t}) - f^\mu_{\phi_k}(w_{i',t'})}^2_2 \right. \\
      &\quad \left. {}+\norm{f^\sigma_{\phi_q}(w_{i,t}) - f^\sigma_{\phi_k}(w_{i',t'})}^2_2 \vphantom{\frac12}\right)
    \end{split}
\end{equation}
We chose the Wasserstein distance because it computes separate distances for the mean and standard deviation. Other metrics like cosine similarity \cite{he2020momentum, chen2020simple} or bilinear product \cite{oord2018representation, henaff2020dataefficient, laskin_srinivas2020curl} do not make this distinction. 

We use the InfoNCE score function \cite{oord2018representation} coupled with our similarity measure to compute the contrastive loss: 
\begin{equation}
    \Loss_{\textrm{TCL}} = -\frac{1}{N} \sum_{i=0}^{N-1}{ \log{\frac{\exp(\textrm{sim}(q_i, k_i))}{\sum_{j=0}^{N-1}{\exp(\textrm{sim}(q_i, k_j))}}}},
\label{eq:contrastive_loss}
\end{equation}
where $k_i$ is the positive key for a given query $q_i$. Given a query and batch of N keys, the InfoNCE loss can be understood as an N-way cross-entropy loss where $k_i$ is the label for the corresponding $q_i$. Algorithm \ref{alg:tcl_contrastive_loss} summarizes the trajectory contrastive loss as pseudo code. 

%\subsubsection{Momentum Encoding}
%We use momentum contrast (MoCo) which was introduced by \cite{he2020momentum}. We use a query encoder $f_{\phi_q}$, parameterized by $\phi_q$, for encoding queries and a key encoder $f_{\phi_k}$, parameterized by $\phi_k$, for encoding keys. Only the query encoder receives gradient from back-propagating the contrastive loss. The key encoder is an exponentially moving average version of the query encoder. Given the query encoder, key encoder, and momentum $m$, the update equation for the key encoder is 
%\begin{equation}
%    \phi_k \leftarrow m \phi_k + (1 - m) \phi_q
%\end{equation}

\subsection{Using TCL in Meta-Training}
TCL is designed to be a plug-and-play component that can be used in any context-based meta-RL algorithm. Consider a generic meta-RL algorithm $\Algo_\theta$ parameterized by $\theta$, query encoder $f_{\phi_q}$, and key encoder $f_{\phi_k}$. The main changes TCL brings to the meta-training procedure of $\Algo_\theta$ are the following: (1) We collect exploration trajectories to be used for context encoding in a replay buffer. (2) We generate query-key encodings from trajectory windows. The queries are shared between $A_\theta$ and TCL for computing their respective losses to be back-propagated to $f_{\phi_q}$. (3) We optimize an auxiliary contrastive loss $\Loss_{\textrm{TCL}}$ in the gradient update to train $f_{\phi_q}$. (4) We update $f_{\phi_k}$ with exponential moving average (EMA). Since TCL is a meta-training method, the meta-testing procedure of $\Algo_\theta$ remains unchanged. We summarize how TCL fits into the meta-training pipeline of $\Algo_\theta$ in Algorithm \ref{alg:tcl_meta_training}. The additional steps for TCL are highlighted. 

% We redefine context as a trajectory window. We train the context encoder with TCL as an auxiliary loss to the base meta-RL algorithm. We summarize the TCL training pipeline is Algorithm \ref{alg:tcl_meta_training}.

\begin{algorithm}[tb]
   \caption{TCL Meta-training \newline \textit{Highlighted lines indicate TCL steps. 
   \newline Details for $\Loss_{\textrm{TCL}}$ are described in Algorithm} \ref{alg:tcl_contrastive_loss}.}
   \label{alg:tcl_meta_training}
\begin{algorithmic}[1]
   \REQUIRE Meta-RL algorithm $\Algo_\theta$, set of training tasks $\Dataset{train} = \{\Tau_i\}$, learning rates $\alpha_1, \alpha_2$, momentum $m$
  \STATE \hl{Initialize context encoder replay buffers $\Beta_i$ for each $\Tau_i$}
   \STATE Initialize $\Algo_\theta$
   \WHILE{not done}
    \FOR{each $\Tau_i$}
     \STATE \hl{Collect trajectories from $\Tau_i$ with policy $\pi_\theta$ and add to $\Beta_i$}
     \STATE Collect data for $\Algo_\theta$
    \ENDFOR
    \FOR{step in training steps}
     \STATE \hl{Initialize query-key sets $\textbf{Q}, \textbf{K}$}
     \FOR{each $\Tau_i$}
      \STATE \hl{Sample context $\{w_{j,t}\} \sim \Beta_i$}
      \STATE \hl{$\textbf{Q}_i, \textbf{K}_i = f_{\phi_q}(w_{j,t}), f_{\phi_k}(w_{j,t'})$}
      \STATE $\Loss^i_{\Algo_\theta} = \Loss_{\Algo_\theta}(\Tau_i)$
     \ENDFOR
     \STATE \hl{$\phi_q \leftarrow \phi_q - \alpha_1 \nabla_{\phi_q} (\Loss_{\textrm{TCL}}(\textbf{Q}, \textbf{K}) + \sum_{i}{\Loss^i_{\Algo_\theta}})$}
     \STATE \hl{$\phi_k \leftarrow m \phi_k + (1 - m) \phi_q$}
     \STATE $\theta \leftarrow \theta - \alpha_2 \nabla_\theta \sum_{i}{\Loss^i_{\Algo_\theta}}$
    \ENDFOR
   \ENDWHILE
\end{algorithmic}
\end{algorithm}

%% file: 5-experiments.tex
\section{Experiments}
\label{sec:exp}

\begin{figure*}
\begin{center}
\includegraphics[width=.8\linewidth]{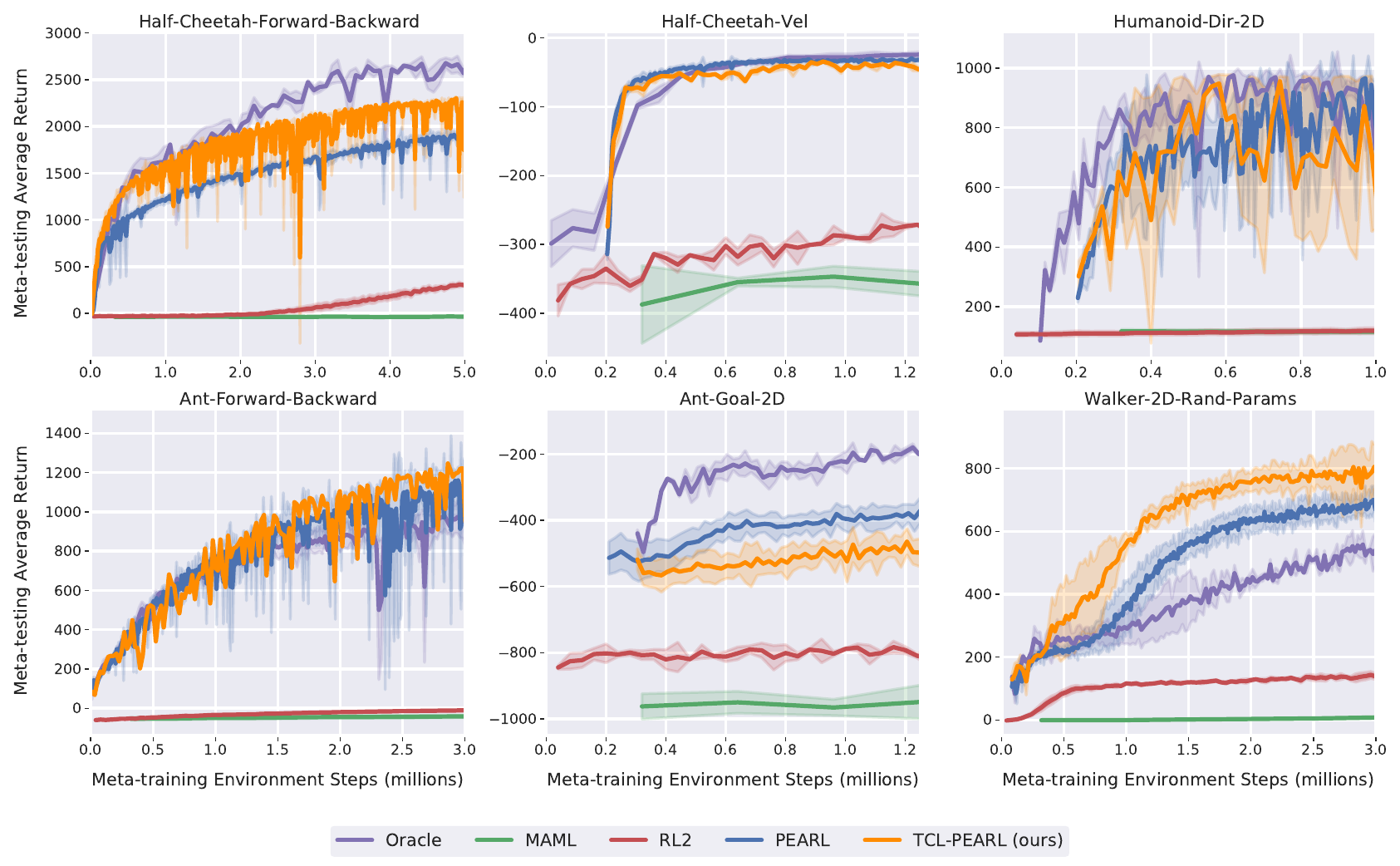}
\end{center}
\vskip -.5cm
  \caption{\textbf{MuJoCo Continuous Controls benchmark: } Test-time average returns vs. environment samples collected during meta-training. Our method performs comparably or better than previous meta-RL methods on 5 of the 6 environments. We calculated the relative performance of TCL-PEARL over PEARL and find that the average of this relative performance gain is 1.13x, and the median is 1.14x. The performance of PEARL on Ant-Goal-2D was reported by \citet{rakelly2019efficient}, but we are not able to reproduce the results. }
\label{fig:mujoco_results}
\vspace{-10pt}
\end{figure*}

In this section, we conduct experiments to better understand the properties of TCL. The questions we hope to answer are the following: (1) Can TCL improve meta-training on a variety of environments?  (2) Can TCL improve a context encoder's representation learning? In order to answer those questions, we perform the following experiments. (1) We test our method on the standard MuJoCo~\citep{todorov2012mujoco} meta-RL benchmark, a commonly used benchmark by previous work~\citep{finn2017modelagnostic,  rothfuss2018promp, rakelly2019efficient, fakoor2020metaqlearning}. (2) We further test our method on the much larger scale Meta-World benchmark~\citep{yu2020meta}. (3)   We plot the context embeddings trained with and without TCL using t-SNE~\citep{vanDerMaaten2008}, and we compare them qualitatively and quantitatively. 

Experimental descriptions, details, and hyperparameters that are important for reproducibility are provided in the appendix. 

\subsection{Experiment Settings}
Meta-RL algorithms are evaluated on how well an agent can distinguish and adapt to unseen test tasks. To evaluate on meta-testing tasks, we provide all methods with the same number of exploration trajectories to adapt their policy to the task at hand. The meta-testing performance is the average returns of trajectories collected after each method has seen the exploration trajectories.

In particular, we evaluated on two sets of environments: a widely-used~\citep{finn2017modelagnostic,  rothfuss2018promp, rakelly2019efficient, fakoor2020metaqlearning} continuous-control meta-RL benchmark simulated via the MuJoCo \cite{todorov2012mujoco} physics engine, which we will now refer to as simply ``MuJoCo Controls", and an object manipulation meta-RL benchmark called Meta-World \cite{yu2020meta}, which is a set of robotic arm manipulation tasks of various complexity (Meta-World is also based on MuJoCo). 

\subsection{Baselines}
In this paper, we implement TCL on top of PEARL~\citep{rakelly2019efficient}. We name this method as TCL-PEARL. Note that the proposed component can be combined with other context-based meta-RL algorithms as well. We compare TCL-PEARL against the PEARL baseline. In addition, we also implement other meta-RL algorithms, including MAML with TRPO \cite{finn2017modelagnostic, schulman2017trust} and RL2 with PPO \cite{duan2016rl2, schulman2017proximal}. Note these algorithms are not directly comparable to TCL-PEARL since they use either different base RL algorithms, or different meta-learning algorithms. 

In order to better understand how well those meta-RL algorithms perform, we establish an oracle performance by training on the goal-conditioned version of the original meta-RL problem. More specifically, instead of learning the context encoders, we provide the ground truth hand-designed context encoder to the agent, and train an RL policy conditioned on this ground truth context encoder with SAC \cite{haarnoja2019soft}. In practice, the reward function and dynamics of a meta-RL environment are parameterized by a task vector that completely characterizes the task at hand. We use this task vector as the context embedding for the oracle. For example, the context embedding for a 2D navigation task would be the Cartesian coordinates of the target location. Walker-2D-Rand-Params in MuJoCo Controls contains 67 system parameters, and we use these parameters to form a context encoder, since they comprehensively described the task. 
Theoretically, this oracle performance serves as an empirical upper bound for the meta-RL algorithm. 
We describe all hyper-parameters in the appendix. 

\subsection{Evaluation on MuJoCo Controls}
\label{exp:mujoco}
The MuJoCo Controls benchmark consists of six continuous control environments focused on robotic locomotion. It was first introduced by \citet{finn2017modelagnostic} and has since been the \textit{de facto} standard for motion tasks. Tasks within a family may differ in reward functions (e.g. walking direction for Humanoid-Dir-2D) or agent dynamics (e.g. random agent system parameters for Walker-2D-Rand-Params). All tasks have a horizon length of 200 steps. All methods are allowed 2 exploration trajectories for meta-testing. 

The meta-testing results for MuJoCo Controls are shown in Figure \ref{fig:mujoco_results}. TCL-PEARL performs comparably or better than the baselines in 5 of the 6 environments, except for the Ant-Goal-2D. Note that we were not able to reproduce PEARL's performance on Ant-Goal-2D based on their implementation. This leads to the worse performance of TCL-PEARL on this task. We calculated the relative performance of TCL-PEARL over PEARL and find that the average of this relative performance gain is 1.13x, and the median is 1.14x.

Perhaps counter-intuitively, in 4 out of 6 environments, PEARL and TCL-PEARL perform better or similarly with the oracle algorithm with hand-crafted context embeddings. There are several explanations for this. Take Walker-2D-Rand-Params as an example, the hand-crafted embedding for this is a 67-dimension vector where each element corresponds to a system parameter. Though it is a comprehensive description of the task, but it can also contain noises and unimportant parameters. Also, the value range of each context embedding vector varies a lot, which leads to difficulty for optimization. These experiments show the importance of learning a better task representation. 

\begin{table}[H]
\begin{center}
\input{tables/metaworld_benchmark_runs.tex}
\end{center}
  \caption{Meta-test returns (in thousands) of TCL-PEARL and PEARL on Meta-World ML1. TCL-PEARL performs better (35 out of 50) or comparably (9 out of 50) in \textbf{44 out of 50} environments. We compute the relative performance of TCL-PEARL over PEARL, the average performance gain is 4.3x, the median is 1.4x.
  }
\label{table:metaworld_benchmark_returns}
\vskip -0.5cm
\end{table}

TCL-PEARL's relative performance gain over PEARL is moderate. Meanwhile, the underwhelming oracle performance reveals that the performance bottleneck for MuJoCo Controls is mainly with the policy instead of task representation. The task distributions of MuJoCo Controls are relatively simple. Specifically, two out of the six environments are binary walking tasks whose task distributions are trivial. Meanwhile, Walker and Humanoid are particularly difficult agents to control, even for single-task benchmarks. Considering these, to better verify the task representation learning, we test TCL on Meta-World, where the task distributions are more complex.

\subsection{Meta-World Evaluations}
\label{sec:meta-world}
Meta-World consists of a family of object manipulation tasks that all share the same robot arm on a table-top setup. The robot arm is a 7-DOF Sawyer robot that is simulated via MuJoCo \cite{todorov2012mujoco}, with the action space corresponding to the velocity of the 3D end-effector and the control of the gripper. We evaluate on the Meta-World ML1 benchmark, which tests the few-shot adaptation to task variation within one environment. A task for a particular environment corresponds to a random initial object and goal position. We follow the same evaluation procedure used for MuJoCo Controls but with different parameters. All tasks have a horizon of 200 steps, and all methods are allowed 10 exploration trajectories per task for adaptation. Each environment has 50 meta-training tasks and 10 held out meta-testing tasks. We limit the number of meta-training environment samples to one million time-steps, as it is sufficient to compare sample efficiency among baselines.

We evaluate TCL-PEARL against the baseline PEARL on 50 Meta-World ML1 environments, ranging from opening a door at random positions (door-open) to placing a puck onto randomly located shelves (shelf-place). We report the test-time performance of TCL-PEARL against the baseline in Table \ref{table:metaworld_benchmark_returns}. The performance is evaluated by the average meta-test returns during the last 100,000 environment steps. TCL-PEARL performs better (35 out of 50) or comparably (9 out of 50) than PEARL in 44 out of 50 environments. Among the remaining 15 environments in which TCL-PEARL does not outperform, we find 9 of them to fall into one of the two cases: (1) both PEARL and TCL-PEARL fail to learn the task distribution, or (2) our performance is at least 90\% of the baseline performance. We calculated the relative performance of TCL-PEARL over PEARL and find that the average of this relative performance gain is 4.3x, and the median is 1.4x. The improvement is much larger than that on MuJoCo Controls.

The larger gain on the Meta-World benchmark can be explained by the increased diversity and complexity of the task distribution. This requires better task representation learning for context-based meta-RL. Since TCL focuses on learning task representations, we see more performance improvements on Meta-World than on MuJoCo Controls.

\begin{table}[h]
\begin{center}
\input{tables/tsne_table}
\end{center}
  \caption{\textbf{t-SNE quantitative study for Figure \ref{fig:tsne}} TCL-PEARL's average distance from each point to its centroid is 26.6,  while PEARL's average distance is 38.8. The smaller intro-centroid distance shows TCL-PEARL's context embedding has clearer clustering, meaning TCL-PEARL is more capable of identifying tasks as a group of tasks. TCL-PEARL's average distance between centroids is 68.5 where PEARL's average distance between centroid is 44.86. The greater inter-centroid distance shows TCL-PEARL is pushing different task's context embedding apart from each other to help distinguish tasks better.}
\label{table:tsne_quantitative_study}
\vskip -0.5cm
\end{table}

\begin{figure*}[h]
\begin{center}
\includegraphics[width=0.9\linewidth]{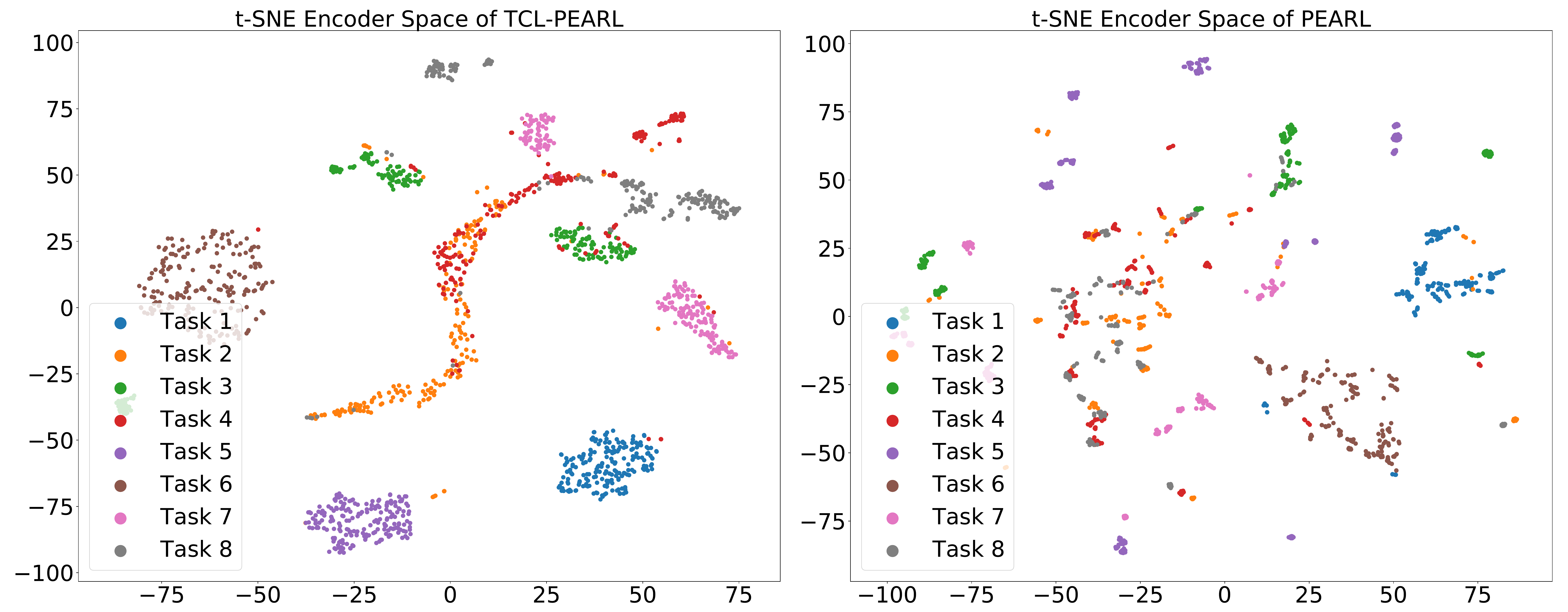}
\end{center}
    \caption{\textbf{t-SNE Context Embedding Space} We prepared a t-SNE plot for TCL-PEARL and PEARL to demonstrate the context embedding space. As shown in the figure, TCL-PEARL's context embedding (left) presents clearer clusters than that of PEARL (right), which are more scattered. In addition, TCL-PEARL's cluster centroids are further apart from each other. This shows TCL forces the context embeddings of the same task closer to each other while pushing those of different tasks apart.}
\label{fig:tsne}
\end{figure*}

\subsection{Learned Context Embedding Visualization}
\label{subsec:tsne}

In order to verify that the improved performance comes from better representation learning, we use t-SNE \citep{vanDerMaaten2008} to visualize the context embeddings trained with and without TCL. 
 
We project the learned context embedding on a 2D plane with the t-SNE method. To collect t-SNE data, we selected 8 distinct tasks under the ``push-v1" Meta-World environment where the agent is encouraged to push an object towards a goal location to receive the reward. The goal locations of these 8 tasks are spread out in a grid, and during each policy rollout we randomized the initial hand positions while keeping the goal and object location fixed. Then we perform 200 policy rollouts for each task to obtain 1,600 trajectories. Finally we sample transition windows on these trajectories and pass the windows to our context encoder to compute the context embedding for the t-SNE plot. We expect that a good context representation should project trajectories corresponding to the same task to closer embedding vectors. Therefore, a better embedding space should exhibit clearer clustering. Our hypothesis is validated by Figure \ref{fig:tsne}, where TCL-PEARL's context encoder shows clearer clustering than PEARL's. Clusters from TCL-PEARL are in general further apart from other clusters, and points within each cluster are closer to each other. 
 
 We also conduct quantitative analysis in Table \ref{table:tsne_quantitative_study}. We measure the average distance of each point to its cluster centroids, and the average distance between cluster centroids. Results show that TCL-PEARL's embedding has smaller intra-cluster distances and larger inter-cluster distances, indicating that each clusters are ``tighter''  and further away from each other. This makes it easier for the conditional policy to distinguish different tasks. These results validate our hypothesis that TCL improves the context encoder training, which translates to better overall performances. 

%% file: tables/metaworld_benchmark_runs.tex
\resizebox{0.4\textwidth}{!}{%
\begin{tabular}{l|c|c}
\toprule
%{} &            PEARL &        TCL-PEARL \\
%Environment               &                  &                  \\
Environments &            PEARL &        TCL-PEARL \\
\midrule
assembly                  &            -0.04 &    \textbf{0.08} \\
basketball                &            -0.07 &   \textbf{-0.02} \\
button-press              &           101.55 &  \textbf{227.21} \\
button-press-topdown      &            32.56 &  \textbf{219.65} \\
coffee-pull               &             0.56 &     \textbf{2.2} \\
coffee-push               &             4.43 &   \textbf{25.85} \\
dial-turn                 &             0.09 &    \textbf{4.96} \\
door-close                &             0.32 &    \textbf{4.69} \\
door-open                 &             2.15 &   \textbf{46.88} \\
door-unlock               &             3.55 &    \textbf{8.96} \\
drawer-close              &            -0.07 &   \textbf{-0.02} \\
drawer-open               &            -0.02 &    \textbf{1.13} \\
faucet-close              &            11.71 &   \textbf{19.37} \\
hammer                    &            -0.04 &   \textbf{-0.02} \\
hand-insert               &           150.64 &  \textbf{241.27} \\
handle-press              &            23.36 &    \textbf{98.4} \\
handle-press-side         &            55.81 &   \textbf{62.28} \\
handle-pull               &            45.28 &    \textbf{46.5} \\
handle-pull-side          &            33.22 &   \textbf{38.38} \\
peg-insert-side           &            -0.03 &    \textbf{0.28} \\
peg-unplug-side           &             2.64 &    \textbf{6.04} \\
pick-place-wall           &             0.11 &    \textbf{0.89} \\
plate-slide               &            26.97 &   \textbf{53.96} \\
plate-slide-back          &             0.22 &    \textbf{3.04} \\
plate-slide-back-side     &             3.12 &    \textbf{7.06} \\
push                      &             4.56 &   \textbf{26.36} \\
push-wall                 &             7.64 &   \textbf{11.57} \\
reach                     &            84.47 &   \textbf{97.33} \\
reach-wall                &           116.06 &  \textbf{134.85} \\
shelf-place               &            -0.06 &    \textbf{3.03} \\
soccer                    &              4.9 &   \textbf{28.85} \\
stick-push                &            -0.03 &    \textbf{0.01} \\
sweep                     &             1.67 &     \textbf{2.4} \\
sweep-into                &            13.05 &   \textbf{18.01} \\
window-close              &            22.09 &   \textbf{91.19} \\
\midrule
bin-picking               &              4.6 &             4.58 \\
box-close                 &                0 &            -0.03 \\
button-press-topdown-wall &           135.58 &           125.25 \\
disassemble               &            -0.02 &            -0.02 \\
lever-pull                &                0 &            -0.01 \\
pick-out-of-hole          &            -0.02 &            -0.02 \\
pick-place                &             2.53 &             0.06 \\
push-back                 &             0.14 &            -0.06 \\
stick-pull                &            -0.04 &            -0.05 \\
\midrule
button-press-wall         &   \textbf{176.5} &           110.39 \\
coffee-button             &  \textbf{157.77} &            94.75 \\
door-lock                 &   \textbf{25.23} &            14.25 \\
faucet-open               &   \textbf{18.81} &            12.83 \\
plate-slide-side          &   \textbf{19.07} &             7.87 \\
window-open               &   \textbf{20.64} &            17.57 \\
\bottomrule
\end{tabular}}

%% file: tables/tsne_table.tex
\resizebox{0.4\textwidth}{!}{%
\begin{tabular}{c|c|c}
\toprule
{}                                  & PEARL & TCL-PEARL \\
\midrule
Average distance to centroid        & 38.8  & 26.6 \\
Average distance between clusters   & 44.86 & 68.5\\
\bottomrule
\end{tabular}}

%% file: 6-conclusion.tex
\section{Conclusion}
In this paper, we propose a contrastive auxiliary loss to improve the training of context-based meta-RL algorithms. The proposed TCL leverages the natural hierachical structure of meta-RL, does not require any labels, and makes no extra assumptions about taks. This makes it widely applicable to context-based meta-RL. Experiments show that TCL significantly improved PEARL, a strong meta-RL baseline, achieving better and similar performances in most of the environments in the MuJoCo Controls benchmark (5 out of 6) and Meta-World (44 out of 50) benchmark. 

%% file: 7-appendix.tex
\newpage
\appendix

\begin{figure*}[ht!]
\begin{center}
\includegraphics[width=0.9\linewidth]{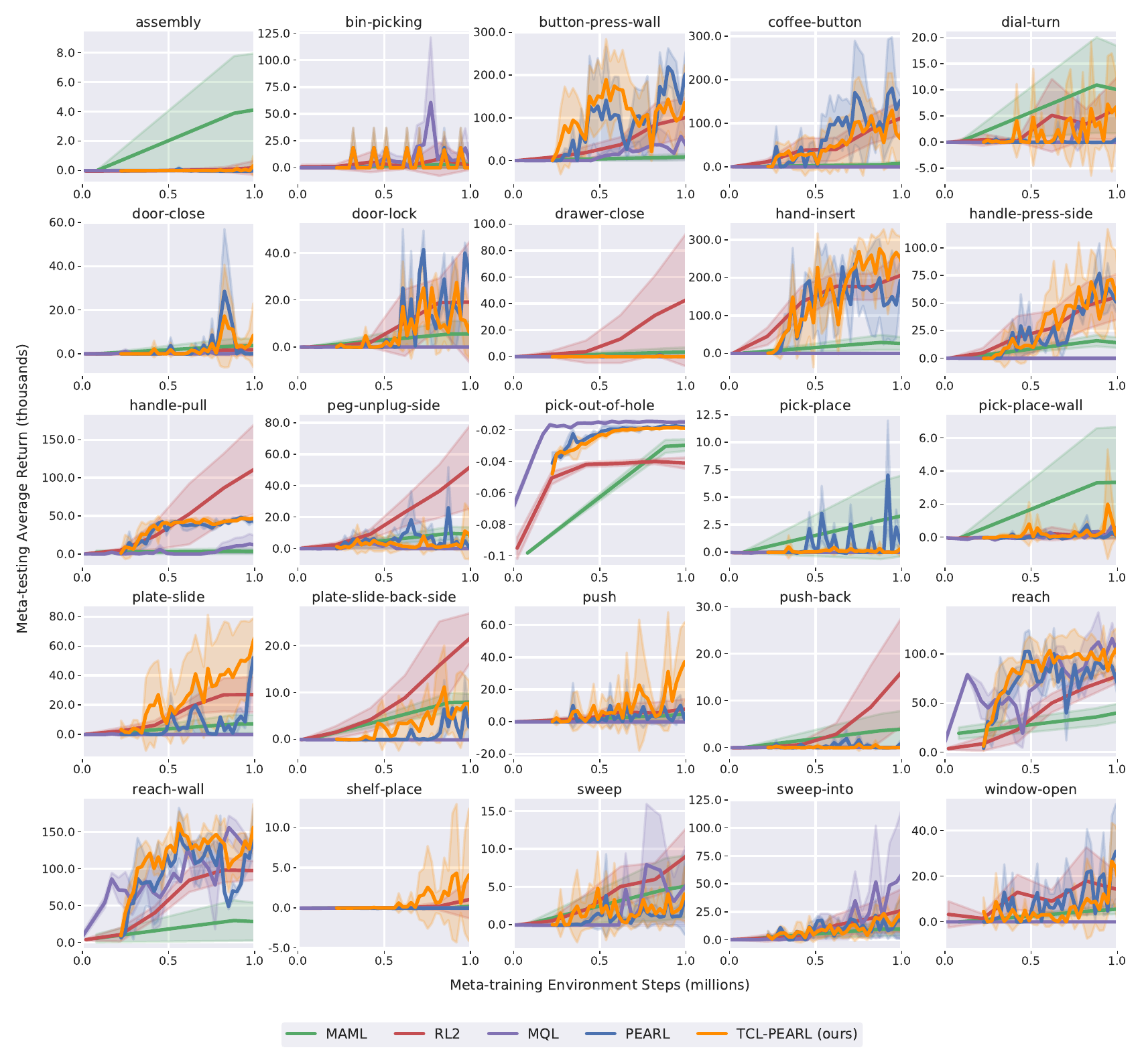}
\end{center}
  \caption{\textbf{Metaworld ML1 Benchmark:} 
  Test-time average returns vs. environment samples collected during meta-training on 25 environments from the Metaworld ML1 benchmark. We meta-train all methods for 1 million environment timesteps. Performance is evaluated by the average test-time returns over the last 100,000 environment timesteps. Our method TCL-PEARL outperforms PEARL in \textbf{17 out of 25} environments and all other existing methods in \textbf{7 out of 25} environments.
  }
\label{fig:metaworld_results}
\end{figure*}

\begin{table*}[ht]
\begin{center}
\input{tables/metaworld_benchmark_runs_full}
\end{center}
  \caption{Meta-test returns (in thousands) and number of ``wins" after 1 million environment steps on Metaworld ML1. The returns are the average of the test-time returns collected over the last 100,000 environment timesteps.  The number of ``wins" is defined as the number of environments in which the algorithm performs better than the rest. TCL-PEARL outperforms the baselines in \textbf{7 out of 25} environments. This is the highest number of ``wins" among all algorithms. 
  }
\label{table:metaworld_benchmark_runs_full}
\end{table*}

\begin{table*}[ht]
\begin{center}
\input{tables/tcl_mujoco_hyperparameters}
\end{center}
  \caption{\textbf{TCL hyper-parameters for MuJoCo Controls. } TCL hyper-parameters are hyper-parameters for tuning the TCL component of TCL-PEARL. The hyper-parameters for tuning the meta-RL component are shared with PEARL. We use the same meta-RL hyper-parameters of PEARL as \citet{rakelly2019efficient}. 
  }
\label{table:mujoco_tcl_hyperparameters}
\end{table*}

\begin{table*}[ht]
\begin{center}
\input{tables/oracle_mujoco_hyperparameters}
\end{center}
  \caption{Oracle context embedding designs for MuJoCo Controls, based on goal-conditioned SAC
  }
\label{table:oracle_mujoco_hyperparameters}
\end{table*}

\begin{table*}[ht]
\begin{center}
\input{tables/mql_metaworld_hyperparameters}
\end{center}
  \caption{MQL hyper-parameters for Meta-World
  }
\label{table:mql_metaworld_hyperparameters}
\end{table*}

\begin{table*}[ht]
\begin{center}
\input{tables/pearl_metaworld_parameters}
\end{center}
  \caption{\textbf{PEARL hyper-parameters for Meta-World.} In our Meta-World experiments, PEARL and TCL-PEARL share the same meta-RL hyper-parameters. Additional TCL-specific hyper-parameters are listed in Table \ref{table:tcl_specific_hyperparameters}
  }
\label{table:pearl_metaworld_hyperparameters}
\end{table*}

\begin{table*}[ht]
\begin{center}
\input{tables/tcl_specific_hyperparameter}
\end{center}
  \caption{\textbf{TCL-specific hyper-parameters of TCL-PEARL for Meta-World.} The hyper-parameters listed are used for the TCL component of TCL-PEARL. In our Meta-World experiments, PEARL and TCL-PEARL share the same meta-RL hyper-parameters. TCL-PEARL has additional hyper-parameters for contrastive learning. This way, we can see the difference brought solely by the additional TCL component built on top of PEARL. }
\label{table:tcl_specific_hyperparameters}
\end{table*}

\begin{table*}[ht]
\begin{center}
\input{tables/metaworld_env_runtime}
\end{center}
  \caption{\textbf{Meta-World environment wall-clock times.} We conducted our experiments on a Linux machines with 8 GPUs and a 16-core CPU. Given the same infrastructure, hyper-parameter, and training iteration, TCL-PEARL takes 10.16 hours to complete whereas its baseline meta-RL algorithm PEARL takes 7.97 hours. }
\label{table:metaworld_benchmark_run_time}
\end{table*}

\section{Experiment Details}
\subsection{Algorithm Evaluation and Comparison}
We trained and evaluated TCL-PEARL against a full suite of baselines (MAML, RL2, MQL, and PEARL) on Meta-World for one million environment timesteps. We report the meta-training performance curves in Figure \ref{fig:metaworld_results}. We also summarize each algorithm's performance by reporting the average meta-test returns during the last 100K environment timesteps in Table \ref{table:metaworld_benchmark_runs_full}. All runs are averaged over three random seeds.

Sample efficiency is an important metric when evaluating meta-RL algorithms. When an agent achieves higher test-time performance with less meta-training environment steps, we conclude that this agent is more sample efficient. For MuJoCo Controls experiments, we followed the same evaluation procedure as originally proposed by  \citet{finn2017modelagnostic}, \citet{rothfuss2018promp}, and \citet{rakelly2019efficient}. For Meta-World experiments, we trained all algorithms up to 1 million environment timesteps, which we empirically found to be a challenging cutoff. To better compare our algorithm against others, we also report the number of ``wins" for a particular algorithm, which we define as the number of environments in which the algorithm performs better than the rest. We report the number of ``wins" for each algorithm on 25 Meta-World ML1 environments in Table \ref{table:metaworld_benchmark_runs_full} based on average returns over the last 100,000 environment timesteps. We found that TCL-PEARL achieved 7 wins, which is the highest among all algorithms. This shows that TCL-PEARL is at least comparable or better in sample efficiency than the state-of-the-art meta-RL algorithms. 

\subsection{Time \& Space Complexity Analysis}
TCL adds an auxiliary loss on top of context-based meta-RL algorithms with minimal space complexity added during data augmentation. The increase in time complexity mostly comes from gradient calculations through the encoder network (which is accelerated by GPU) and data augmentation including trajectory windowing (computed by CPU). The later part consumes more computing resources as this step is applied to every training iteration. We also compared TCL-PEARL's real world run-time against PEARL's and report our findings in Table \ref{table:metaworld_benchmark_run_time} given the same set of hyper-parameters and number of training iterations. We conducted our experiments mostly on a Linux machine with 8 Tesla M40 GPUs and a 16 core CPU with 32 threads in total. Completing Meta-World meta-training with one million environment steps takes 6.75 hours to complete whereas its base meta-RL algorithm PEARL takes 5.17 hours.

\section{Meta-RL Benchmark Details}
\subsection{MuJoCo Controls Details}
We used the MuJoCo Controls benchmark from \url{https://github.com/katerakelly/oyster}, which is open-sourced by \citet{rakelly2019efficient}. This benchmark is a collection of meta-RL environments that were first open-sourced by \citet{finn2017modelagnostic} and \citet{rothfuss2018promp}. We used the same train/test task splits for MuJoCo Controls as \citet{rakelly2019efficient}. 

\subsection{Meta-World Details}
We used the Meta-World codebase from \url{https://github.com/rlworkgroup/metaworld}, which is open-sourced by \citet{yu2020meta}. In our Meta-World experiments, we used the ML1 benchmark following the same train/test task splits provided by \citet{yu2020meta}. Specifically, during the algorithm initialization, we sample 50 train tasks in $\Dataset{train}$ and 10 test tasks in $\Dataset{test}$. Each task is characterized by a goal position and initial object position. 

\section{Implementation Details \& Hyper-parameters}
Since TCL builds on top of context-based meta-RL and minimally modifies the algorithm, TCL-PEARL shares the same meta-RL hyper-parameters with PEARL for all our experiments. The hyper-parameters unique to TCL-PEARL are those used for tuning contrastive learning. We refer to these contrastive learning hyper-parameters as ``TCL hyper-parameters" and the rest of the hyper-parameters to be ``meta-RL hyper-parameters".

\subsection{Implementation for MuJoCo Controls}
We used a 2-layer MLP with hidden size of 200 as a context encoder, context dimension of 5, and momentum $0.005$ for all environments. We report the environment-specific TCL hyper-parameters for MuJoCo Controls in Table \ref{table:mujoco_tcl_hyperparameters}. We used the hyper-parameters for MAML, RL2, and PEARL proposed by \citet{rakelly2019efficient}. 

The oracle we used in the MuJoCo Controls experiments is based on the multi-task SAC, which adapts SAC to the multi-task setting by appending the one-hot task ID to the input. This method works well for multi-task RL but is not suitable for meta-RL because in multi-task setting, the test tasks must come from the train tasks. In order to adopt this method for our experiments, we modified multi-task SAC such that instead of one-hot task encodings, we use hand-crafted context embeddings. In practice, the reward function and dynamics of a meta-RL environment are parameterized by a task vector that completely characterizes the task at hand. We use this task vector as the context embedding for the oracle. We use the same hyper-parameters as multi-task SAC used by \citet{yu2020meta}. The only difference is in the context embedding design, which we report in Table \ref{table:oracle_mujoco_hyperparameters}. We base our oracle implementation on the multi-task SAC provided by Garage.

\subsection{Implementation for Meta-World}
Large-scale experiments on Meta-World environments are hard to fine-tune because each environment reacts dramatically differently to a specific hyper-parameter tuning. We explored several different sets of hyper-parameters including number of exploration steps, network sizes, context embedding dimensions, batch sizes, replay buffer sizes, \textit{etc}. One set of hyper-parameters that works well on one environment often shows detrimental effect on other environments. Due to computational constraints, we chose to use the best known set of hyper-parameters for PEARL and then used them for TCL-PEARL. We report the hyper-parameters shared by PEARL and TCL-PEARL in Table \ref{table:pearl_metaworld_hyperparameters}. For TCL-specific hyper-parameters for Meta-World, see Table \ref{table:tcl_specific_hyperparameters}.

We used the same hyper-parameters for MAML and RL2 as those proposed by \citet{yu2020meta}. For the Meta-World experiments, we adopted implementations for MAML, RL2, and PEARL from the open-sourced RL toolkit Garage, which can be found at \url{https://github.com/rlworkgroup/garage}. TCL-PEARL was built on top of the Garage implementation of PEARL. We used the Meta-Q-Learning codebase from \url{https://github.com/amazon-research/meta-q-learning} which was open-sourced by \citet{fakoor2020metaqlearning}. We report the hyper-parameters for MQL on Meta-World in Table \ref{table:mql_metaworld_hyperparameters}. 

\subsection{Data Augmentation}
We explored two different data augmentation strategies: (1) sample random transitions from the trajectory, and (2) crop a window of transitions. Empirically we found that the window cropping strategy often provides more local trajectory shaping information. This helps the context encoder to better distinguish a trajectory window from others which leads to better performance. We provide a python-like pseudocode below to help explain how to perform this data augmentation:

\vspace{5mm}
\begin{lstlisting}[language=Python]
from numpy.random import randint
# Note: window_size < max_traj_len
def window_crop(traj, window_size, max_traj_len):
    window = {}
    bound = max_traj_len - window_size
    start = randint(0, bound)
    end = start + window_size
    window['obs'] = traj['obs'][start: end]
    window['a'] = traj['a'][start: end]
    window['r'] = traj['r'][start: end]
    return window
\end{lstlisting}

% \begin{minted}{python}
% from numpy.random import randint
% # Note: window_size < max_traj_len
% def window_crop(traj, window_size, max_traj_len):
%     window = {}
%     bound = max_traj_len - window_size
%     start = randint(0, bound)
%     end = start + window_size
%     window['obs'] = traj['obs'][start: end]
%     window['a'] = traj['a'][start: end]
%     window['r'] = traj['r'][start: end]
%     return window
% \end{minted}

\subsection{Data Collection Through Policy Rollouts}
During each training iteration, the agent is allowed to explore the task environment for $N_{\textrm{prior}}$ rollouts with context embedding $\textbf{z}_{\textrm{prior}}$ sampled from the unit normal prior $\Normal(\textbf{0}, \textbf{1})$. Then we update the agent's belief by generating a new context embedding $\textbf{z}_{\textrm{posterior}}$ of a windowed trajectory sampled from a replay buffer. We perform $N_{\textrm{posterior}}$ policy rollouts conditioned on context embeddings inferred from past exploration trajectories. The reason to have policy rollouts with $\textbf{z}_{\textrm{prior}}$ is to make sure the agent explores the environment well enough so that it does not converge to a sub-optimal strategy, whereas the reason for having policy rollouts with $\textbf{z}_{
\textrm{posterior}}$ is to make sure that the agent uses its best knowledge to complete the task.

%% file: tables/metaworld_benchmark_runs_full.tex
\begin{tabular}{l|ccccc}
\toprule
Environment &            MAML &             RL2 &              MQL &            PEARL &        TCL-PEARL \\

\midrule
assembly              &   \textbf{3.88} &            0.05 &            -0.01 &            -0.04 &             0.08 \\
bin-picking           &            3.12 &            7.95 &   \textbf{12.71} &              4.6 &             4.58 \\
button-press-wall     &            6.87 &           83.44 &            36.71 &   \textbf{176.5} &           110.39 \\
coffee-button         &            4.62 &           85.65 &            -0.08 &  \textbf{157.77} &            94.75 \\
dial-turn             &  \textbf{10.92} &            3.46 &            -0.07 &             0.09 &             4.96 \\
door-close            &            3.44 &            1.62 &              0.4 &             0.32 &    \textbf{4.69} \\
door-lock             &            5.53 &           18.87 &            -0.04 &   \textbf{25.23} &            14.25 \\
drawer-close          &            3.15 &  \textbf{31.13} &            -0.04 &            -0.07 &            -0.02 \\
hand-insert           &           28.77 &          175.83 &            -0.06 &           150.64 &  \textbf{241.27} \\
handle-press-side     &            15.8 &           47.41 &            -0.09 &            55.81 &   \textbf{62.28} \\
handle-pull           &            3.57 &  \textbf{86.63} &            12.04 &            45.28 &             46.5 \\
peg-unplug-side       &            9.65 &  \textbf{36.51} &            -0.02 &             2.64 &             6.04 \\
pick-out-of-hole      &           -0.03 &           -0.04 &   \textbf{-0.02} &            -0.02 &            -0.02 \\
pick-place            &   \textbf{2.89} &           -0.03 &            -0.02 &             2.53 &             0.06 \\
pick-place-wall       &   \textbf{3.29} &            0.09 &             0.36 &             0.11 &             0.89 \\
plate-slide           &            6.97 &           26.92 &             -0.1 &            26.97 &   \textbf{53.96} \\
plate-slide-back-side &            7.86 &  \textbf{15.81} &            -0.11 &             3.12 &             7.06 \\
push                  &            3.76 &            5.41 &             -0.1 &             4.56 &   \textbf{26.36} \\
push-back             &            3.57 &   \textbf{8.56} &            -0.02 &             0.14 &            -0.06 \\
reach                 &           35.91 &           66.06 &  \textbf{106.42} &            84.47 &            97.33 \\
reach-wall            &           29.79 &           98.21 &           129.91 &           116.06 &  \textbf{134.85} \\
shelf-place           &           -0.01 &             0.2 &            -0.02 &            -0.06 &    \textbf{3.03} \\
sweep                 &            4.65 &   \textbf{5.96} &              3.7 &             1.67 &              2.4 \\
sweep-into            &            9.48 &           19.18 &   \textbf{50.13} &            13.05 &            18.01 \\
window-open           &            4.91 &           17.87 &            -0.04 &   \textbf{20.64} &            17.57 \\
\midrule
\textbf{Number of wins}           &   4 &               6 &                4 &                4 &       \textbf{7} \\
\bottomrule
\end{tabular}

%% file: tables/tcl_mujoco_hyperparameters.tex
\begin{tabular}{l|cccccc}
\toprule
{}                      & Cheetah-FB & Cheetah-Vel & Humanoid & Ant-FB & Ant-Goal-2D & Walker \\
\midrule
Window size             &        128 &          64 &      128 &     64 &         128 &     64 \\
% context dimension       &          5 &           5 &        5 &      5 &           5 &      5 \\
Contrastive loss scale  &          1 &           1 &        5 &      1 &           1 &      5 \\
% momentum                &      0.005 &       0.005 &    0.005 &  0.005 &       0.005 &  0.005 \\
\bottomrule
\end{tabular}

%% file: tables/oracle_mujoco_hyperparameters.tex
\begin{tabular}{l|cccccc}
\toprule
{}                           &     Cheetah-FB &   Cheetah-Vel &          Humanoid &         Ant-FB &      Ant-Goal-2D & Walker \\
\midrule
Context embedding            & one-hot goal dir & goal velocity & goal dir & one-hot goal dir & goal position &   agent params \\
Dimensions &              2 &             1 &                 2 &              2 &                2 &      67 \\
\bottomrule
\end{tabular}

%% file: tables/mql_metaworld_hyperparameters.tex
\begin{tabular}{l|c}
\toprule
Hyper-parameter                                                      &   Hyper-parameter value \\
\midrule
$\beta$ clipping                                                     &   1.1 \\
TD3 exploration noise                                                &   0.2 \\
TD3 policy noise                                                     &   0.2 \\
TD3 policy update frequency                                          &     3 \\ 
Parameter updates per meta-training iteration                        &   200 \\
Vanilla off-policy adaptation updates per episode                    &    10 \\
Importance-ratio corrected off-policy adaptation updates per episode &   400 \\
GRU sequence length                                                  &    10 \\
Context dimension                                                    &    30 \\
Adam learning rate                                                   & 0.008 \\
\bottomrule
\end{tabular}

%% file: tables/pearl_metaworld_parameters.tex
% \resizebox{\textwidth}{!}{
% \begin{tabular}{c|c|c}
% \toprule
% Hyper-parameters                & TCL-PEARL & PEARL\\
% \midrule
% num-steps-per-epoch             & 4000      & 4000 \\
% num-initial-steps               & 4000      & 4000 \\
% num-steps-prior                 & 800       & 800 \\
% num-extra-rl-steps-posterior    & 800       & 800 \\
% num-epochs                      & 50        & 50 \\
% num-train-tasks                 & 50        & 50 \\
% num-test-tasks                  & 10        & 10 \\
% latent-size                     & 7         & 7 \\
% encoder-hidden-size             & 400       & 400 \\
% net-size                        & 400       & 400 \\
% batch-size                      & 256       & 256 \\
% embedding-batch-size            & 128       & 128 \\
% embedding-mini-batch-size       & 128       & 128 \\
% meta-batch-size                 & 16        & 16 \\
% num-tasks-sample                & 15        & 15 \\
% reward-scale                    & 10        & 10 \\
% max-path-length                 & 200       & 200 \\
% replay-buffer-size              & 1000000   & 1000000 \\
% \bottomrule
% \end{tabular}%}
\begin{tabular}{l|c}
\toprule
Hyper-parameter                & Hyper-parameter value \\
\midrule
Parameter updates per epoch            & 4000      \\
Warmup timesteps per task               & 4000      \\
Exploration timesteps per meta-training task                 & 800       \\
Posterior timesteps per meta-training task    & 800       \\
Epochs                  & 50        \\
Context embedding dimension                    & 7         \\
Context encoder hidden size             & 400       \\
Policy hidden size                        & 400       \\
Batch size                      & 256       \\
Context embedding batch size            & 128       \\
Meta-batch size                 & 16        \\
Training tasks sampled per epoch                & 15        \\
Reward scale                   & 10        \\
Replay buffer size             & 1000000   \\
\bottomrule
\end{tabular}%}

%% file: tables/tcl_specific_hyperparameter.tex
\begin{tabular}{l|c}
\toprule
Hyper-parameter                                & Hyper-parameter value \\
\midrule
Trajectory Augmentation Window Size            & 64      \\
Contrastive Loss Temperature                   & 1         \\
\bottomrule
\end{tabular}

%% file: tables/metaworld_env_runtime.tex
\resizebox{\textwidth}{!}{
\begin{tabular}{l|c|c|l|c|c}
\toprule
Meta-World                  & TCL-PEARL                   & PEARL                       & Meta-World                    & TCL-PEARL             & PEARL  \\
Environment Name            & Run-Time                    & Run-Time                    & Environment Name              & Run-Time              &Run-Time\\
\midrule
faucet-open-v1              & 24363.49 s                  & 18181.62 s                  & bin-picking-v1                & 24212.85 s            & 18203.74 s\\
sweep-v1                    & 24513.58 s                  & 18591.64 s                  & plate-slide-back-v1           & 23520.31 s            & 18195.49 s\\
basketball-v1               & 24414.36 s                  & 19035.51 s                  & drawer-close-v1               & 24515.42 s            & 18041.66 s\\
sweep-into-v1               & 24947.83 s                  & 19074.42 s                  & button-press-topdown-v1       & 23864.72 s            & 17992.37 s\\
faucet-close-v1             & 24492.32 s                  & 18871.45 s                  & reach-v1                      & 24477.98 s            & 18205.53 s\\
push-back-v1                & 24999.32 s                  & 19346.44 s                  & button-press-topdown-wall-v1  & 24062.87 s            & 18104.58 s\\
lever-pull-v1               & 23997.50 s                  & 19459.97 s                  & reach-wall-v1                 & 25201.47 s            & 18960.29 s\\
dial-turn-v1                & 23480.95 s                  & 18469.58 s                  & peg-insert-side-v1            & 24954.97 s            & 18116.93 s\\
stick-push-v1               & 24380.38 s                  & 18279.03 s                  & push-v1                       & 24089.16 s            & 18259.77 s\\
coffee-button-v1            & 24444.54 s                  & 18703.70 s                  & push-wall-v1                  & 24218.12 s            & 19611.21 s\\
handle-pull-side-v1         & 23911.76 s                  & 18658.58 s                  & pick-place-wall-v1            & 24119.04 s            & 18791.37 s\\
assembly-v1                 & 24792.06 s                  & 18872.15 s                  & button-press-v1               & 23843.74 s            & 17976.65 s\\
stick-pull-v1               & 24464.69 s                  & 19045.46 s                  & pick-place-v1                 & 23655.13 s            & 17922.44 s\\
pick-out-of-hole-v1         & 24201.22 s                  & 19256.38 s                  & coffee-pull-v1                & 23564.80 s            & 18391.25 s\\
disassemble-v1              & 24944.64 s                  & 19028.64 s                  & peg-unplug-side-v1            & 24242.81 s            & 19249.35 s\\
shelf-place-v1              & 24641.35 s                  & 18614.48 s                  & window-close-v1               & 23726.63 s            & 18398.25 s\\
coffee-push-v1              & 23921.07 s                  & 18244.83 s                  & window-open-v1                & 24348.17 s            & 18687.93 s\\
handle-press-side-v1        & 25002.12 s                  & 18983.30 s                  & door-open-v1                  & 23984.81 s            & 18324.07 s\\
hammer-v1                   & 23696.70 s                  & 18429.84 s                  & door-close-v1                 & 24250.81 s            & 18495.95 s\\
plate-slide-v1              & 24481.81 s                  & 18302.40 s                  & drawer-open-v1                & 24683.07 s            & 19664.53 s\\
plate-slide-side-v1         & 24217.56 s                  & 19471.69 s                  & hand-insert-v1                & 24864.39 s            & 18720.50 s\\
button-press-wall-v1        & 24751.90 s                  & 17877.36 s                  & box-close-v1                  & 24454.36 s            & 18305.75 s\\
handle-press-v1             & 23194.78 s                  & 18200.38 s                  & door-lock-v1                  & 23717.05 s            & 18359.74 s\\
handle-pull-v1              & 24416.05 s                  & 17975.80 s                  & door-unlock-v1                & 24986.74 s            & 19366.95 s\\
soccer-v1                   & 24619.45 s                  & 18280.78 s                  &                               &                       &           \\
plate-slide-back-side-v1    & 24445.71 s                  & 19587.77 s                  & \textbf{Average Run Time}     & \textbf{24305.93 s}   & \textbf{18623.79 s}\\
\bottomrule
\end{tabular}}